\documentclass[conference]{IEEEtran}
\IEEEoverridecommandlockouts
\usepackage[letterpaper, left=0.57in, right=0.57in, bottom=0.57in, top=0.57in]{geometry}
\usepackage{cite}
\usepackage{amsmath,amssymb,amsfonts}
\usepackage{caption, copyrightbox, subfig}
\usepackage{algorithm}
\usepackage{algpseudocode}
\usepackage{graphicx}
\usepackage{textcomp}
\usepackage{multirow}
\usepackage{xcolor}
\usepackage{url}
\usepackage[hidelinks]{hyperref}
\def\BibTeX{{\rm B\kern-.05em{\sc i\kern-.025em b}\kern-.08em
    T\kern-.1667em\lower.7ex\hbox{E}\kern-.125emX}}
\begin{document}

\title{Predicting Drug Effects from High-Dimensional, Asymmetric Drug Datasets by Using Graph Neural Networks: A Comprehensive Analysis of Multitarget Drug Effect Prediction
\thanks{This manuscript has been authored by UT-Battelle LLC under contract DE-AC05-00OR22725 with the US Department of Energy (DOE). The US government retains and the publisher, by accepting the article for publication, acknowledges that the US government retains a nonexclusive, paid-up, irrevocable, worldwide license to publish or reproduce the published form of this manuscript or allow others to do so, for US government purposes. DOE will provide public access to these results of federally sponsored research in accordance with the DOE Public Access Plan(\url{https://www.energy.gov/doe-public-access-plan}).}
}

\author{\IEEEauthorblockN{Avishek Bose}
\IEEEauthorblockA{\textit{Learning Systems Group, Data and AI Systems Section,} \\
\textit{Computer Science and Mathematics Division,}\\
\textit{Oak Ridge National Laboratory}\\
Oak Ridge, TN, USA\\
bosea@ornl.gov}
\and
\IEEEauthorblockN{Guojing Cong}
\IEEEauthorblockA{\textit{Learning Systems Group, Data and AI Systems Section,} \\
\textit{Computer Science and Mathematics Division,}\\
\textit{Oak Ridge National Laboratory}\\
Oak Ridge, TN, USA\\
congg@ornl.gov}
}

\IEEEaftertitletext{\vspace{-2\baselineskip}}

\maketitle

\begin{abstract}
  Graph neural networks (GNNs) have emerged as one of the most effective ML techniques for drug effect prediction from drug molecular graphs. Despite having immense potential, GNN models lack performance when using datasets that contain high-dimensional, asymmetrically co-occurrent drug effects as targets with complex correlations between them. Training individual learning models for each drug effect and incorporating every prediction result for a wide spectrum of drug effects are impractical. Therefore, an opportunity exists to address this challenge as multitarget prediction problems and predict all drug effects at a time. We developed standard and hybrid GNNs to perform two separate tasks: multiregression for continuous values and multilabel classification for categorical values contained in our datasets. Because multilabel classification makes the target data even more sparse and introduces asymmetric label co-occurrence, learning these models becomes difficult and heavily impacts the GNN's performance. To address these challenges, we propose a new data oversampling technique to improve multilabel classification performances on all the given imbalanced molecular graph datasets. Using the technique, we improve the data imbalance ratio of the drug effects while protecting the datasets' integrity. Finally, we evaluate the multilabel classification performance of the best-performing hybrid GNN model on all the oversampled datasets obtained from the proposed oversampling technique. In all the evaluation metrics (i.e., precision, recall, and F1 score), this model significantly outperforms other ML models, including GNN models when they are trained on the original datasets or oversampled datasets with MLSMOTE, which is a well-known oversampling technique.
\end{abstract}


\section{Introduction}\label{sec:inro}
A tangible understanding of unknown drug responses in the human body has always been elusive during the drug development phase \cite{likeness1, lit_database2}. Modeling drug effects before physical trials is indispensable in the drug discovery domain but poses various computational challenges. The procedure of representing drug molecular structures as graphs and learning these graphs by using graph neural networks (GNNs) for drug effect prediction has become the most effective AI/ML-driven pharmacological approach \cite{graphgnn1, graphgnn2} yet to be fully incorporated into the drug design process. This study focuses on drug effect prediction \cite{lit_database3, toxicity2}. This problem is similar to that of popular drug property prediction from the perspective of molecular representation learning but more challenging for ML models \cite{deep2, deep3} due to learning from asymmetrically distributed effects from the datasets as targets.

 Many real-world drug datasets restricted for open access include physical trial effects on the human body, and the number of effects ranges from a few hundred to over a few thousand for each drug. Training an individual model for each drug effect prediction and combining all the model's results while considering the drug effects' co-occurrences are impractical. For example, building a model for predicting a particular drug effect would not account for other co-occurring effects of the same drug. Therefore, combining prediction results of different models of the same drug would be inappropriate for providing a generic notion of the drug effects. 

The datasets used in this study contain drug molecular graphs mapped to their responses (effects) to the human body, and the effects include both categorical values and continuous values as targets. Realizing the prediction objective, we formulate two separate tasks: (1) building GNN-based multiregression models for continuous values and (2) building GNN-based multilabel classification models for categorical values to support effective experimentation of the ML models and simple demonstration of their performances. Although the datasets used in this study are not publicly available, Section \ref{subsec:dataimbal} provides an explanation and a demonstration of each dataset distribution.

In drug effect prediction, even a rarely observed phenomenon (positive drug effect) is significant due to its potential impact on human bodies. However, the seemingly rare minority labels of categorical targets in these datasets cause immense data sparsity, thereby creating an asymmetric data distribution and impeding the learning of GNN models for multitarget prediction. Despite this challenging task, we need GNN models for multilabel classification that are representative and that can effectively model drug effects when trained on high-dimensional, asymmetric datasets. Strategies such as oversampling, downsampling, or synthetic data generation that help ML models adapt to highly asymmetric data distribution demonstrate limited impact on improving model performance.

In this work, we implemented standard and hybrid GNN models from a set of widely used GNN layers: a graph convolutional network (GCN) \cite{gcn}, a graph attention network (GAT) \cite{gat}, a Gaussian mixture model (GMM) \cite{gmm}, and AttentiveFP \cite{attentiveFP}. 
We conducted an extensive experimental analysis to evaluate the GNN's modeling performances on high-dimensional, asymmetrically co-occurrent datasets and discussed their learning limitations.
The standard GNN models take only molecular graphs as inputs. By contrast, the hybrid GNN models take molecular graphs with molecules'  fingerprints as inputs. We used the atom-pair fingerprint \cite{atompair_finger} for the hybrid GNN models because it demonstrates better performance than other fingerprints.  

To address high data imbalance and asymmetric label co-occurrence for multilabel classification tasks using GNN models, we maintain a balanced data distribution while maintaining optimal data integrity. For this reason, we introduce an oversampling technique in \textbf{Algorithm} \ref{alg:heurisic} for oversampling instances of underrepresentative labels (drug effects) while minimizing co-occurrence between them. It should be noted that we use three terms---\textit{labels}, \textit{targets}, and \textit{effects}---interchangeably throughout the study. Finally, we provide guidelines to improve performance in predicting multiple drug effects simultaneously on highly imbalanced multilabel datasets. The contributions of this study are as follows:
    \begin{enumerate}
        \item We implemented a set of standard and hybrid GNNs for multilabel regression and multilabel classification tasks (GNNs) and proposed the best-performing hybrid GNN architecture.
        \item We propose a new data oversampling for the GNN multilabel classification models technique while optimally impacting the datasets' integrity 
        \item We present the complexity of the target space of our graph datasets by reporting the scores of data imbalance ratio metrics IRLbl and SCUMBLE.
        \item We show that oversampled datasets obtained from the proposed oversampling technique demonstrate better imbalance metrics scores compared with MLSMOTE.
        \item We evaluate the multilabel classification performance of the best-performing hybrid GNN model, which improves the low prediction performance on the oversampled datasets and outperforms all previous prediction models for multilabel classification.
    \end{enumerate}

\section{Previous Work}
An ever-growing number of newly discovered drugs and the amount of physicochemical and pharmacokinetic data they generate during their biochemical synthesis processes are beyond the analyzing capability of traditional computing approaches \cite{lit_database2, lit_database3}. Such approaches that leverage empirical ``drug-likeness'' rules \cite{likeness1, likeness2} and associated reasoning \cite{pattern_descriptor} for chemical (sub) structures have become incompatible in addressing the current challenges and specific needs associated with the domain of drug discovery. However, recent advancements in AI/ML technologies \cite{deep2, deep3, deep4} have been reducing the highly expensive lab experimentation costs and time. These advancements have also started revolutionizing the drug discovery domain in various aspects spanning from drug property prediction such as predicting metabolism, absorption, and toxicity to drug effect prediction for humans and human bodies\cite{ADMT1, toxicity1, toxicity2}. From the perspective of deep learning, three major sources of distinct information exist for encoding drug molecular representation: (1) descriptor information \cite{likeness1, pattern_descriptor}, (2) graph structure (atom-atom bond topology) \cite{graphgnn2, attentiveFP}, and (3) geometry (e.g., 2D/3D conformation) \cite{geomgcl, geognn}. Although adopting 3D conformation in molecular representation showed much potential, a list of major shortcomings such as alignment invariance,  calculation cost, and uncertain conformation generation has provided a limited scope to expand. However, the graph-based molecular representation that considers atoms as nodes and atom bonds as edges is more stable and represents complex electronic structures. This representation captures the key interaction between electrons and nuclei in a molecule implicitly as an encoding of its geometry and property.

Traditional ML algorithms such as linear regression, random forest, support vector machine, and XGBoost using molecular descriptors or fingerprints have become incompatible in evolving drug discovery research due to their dependence on feature engineering \cite{LR2, svm2, RF1}. Since this molecular information has been utilized, experimented with, and validated for a long time, we can still benefit from it by using it alongside graph-based molecular representation. An emerging deep architecture called GNN \cite{gnn} uses graph convolution to learn from graph-structured molecular data and encode it into a low-dimensional feature representation. In our work, we focus on GNN-based approaches \cite{couldGNN, cheminet} to learn from the molecular graphs because they showed better performance than traditional ML approaches.

\section{Methodologies and Experiments}
In this section, we first describe a generic GNN architecture for multiregression and multilabel classification tasks. Then, we describe the features and associated challenges of the studied datasets. Subsequently, we introduce an oversampling technique to address these challenges. Finally, we compare this technique with that of Charte et al. \cite{mlsmote} by using five datasets.

\subsection{Proposed Hybrid GNN Architecture}
\begin{figure*}
\centering
\includegraphics[width=0.70\textwidth]{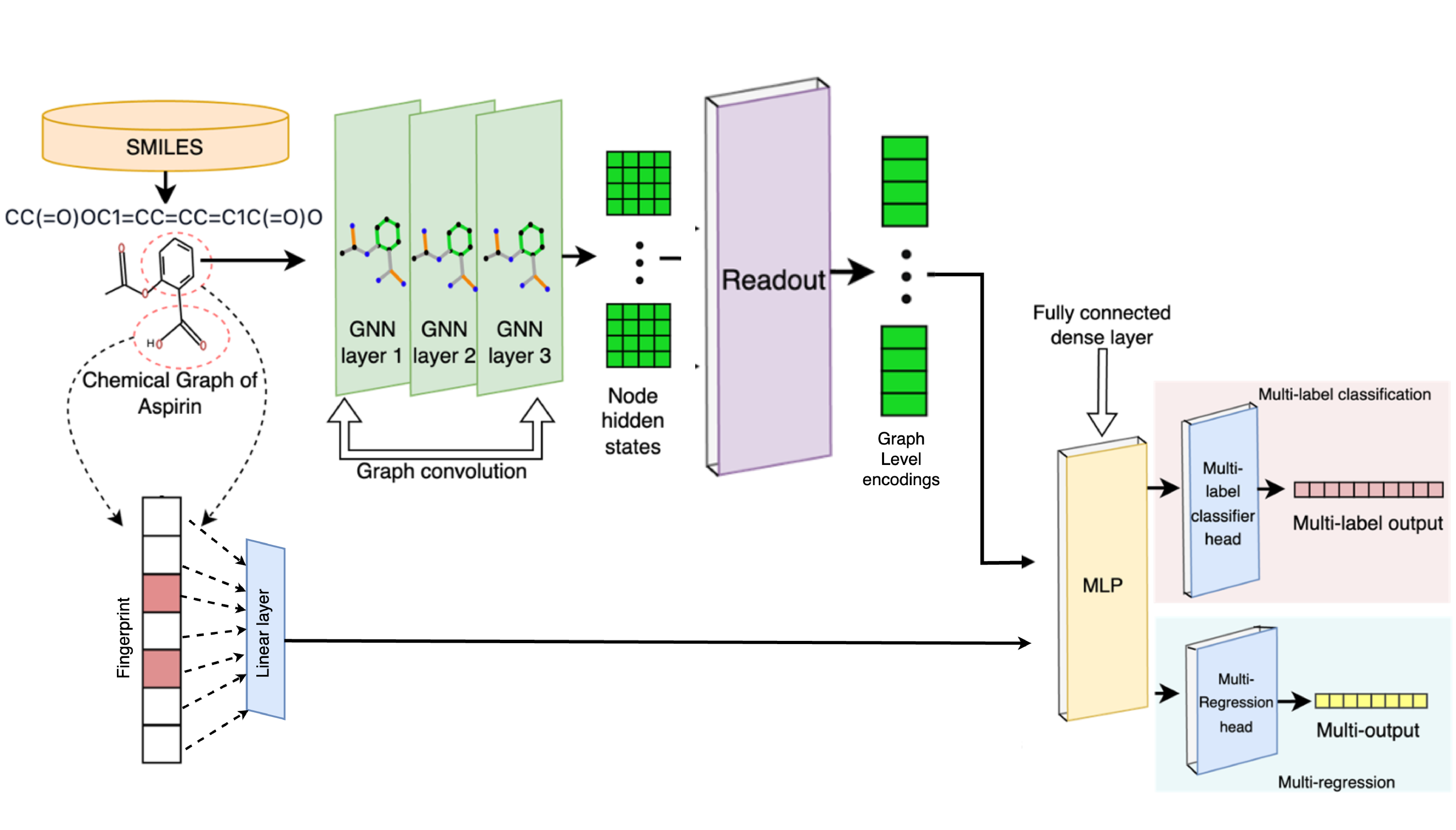}
\caption{Generic network architecture of implemented hybrid GNN models for multiregression and multilabel classification.}
\label{fig:gnn_arch}
\end{figure*}

We propose a generic hybrid GNN architecture shown in \textbf{Figure} \ref{fig:gnn_arch} for both multiregression and multilabel classification tasks in drug effect prediction. For simplicity, we display a single architecture for the two tasks, but they are constructed separately. We can observe from the figure that both the simplified molecular-input line-entry system (SMILES) string and fingerprint are given as inputs to the model. The SMILES strings are transformed into graph inputs using the PCQM4Mv2 package \cite{ogb_bench}, and fingerprints are collected from the rdkit package \cite{atompair_finger}. The GNN layers in the figure are selected from any of the four considered layers---GCN, GAT, GMM, and AttentiveFP---at a time for building the hybrid GNN. The GNN part of the hybrid network produces node embedding (which is atom embedding in our case), which is later fed into the readout layer for generating graph-level embedding (drug molecular embedding). By contrast, the standard GNN architectures do not use fingerprint inputs for multi-output prediction.

To understand the hybrid network in depth, we consider the proposed generic hybrid GNNs trained on drug molecular graphs in which each molecular graph \( G \) consists of multiple atoms \( V \) (\( \{v_1, v_2, \ldots, v_n\} \)) and bonds between atoms \( E \) (\(\{(v_i, v_j) \mid v_i, v_j \in V\} \)). Each atom \( v_i \in V \) has associated feature vector \( \mathbf{x}_i \in \mathbb{R}^d \), where \( d \) is the dimensionality of atom feature space and of all node feature matrix \( \mathbf{X} \in \mathbb{R}^{n \times d} \). The PCQM4Mv2 package provides a total of 9 features for each atom (node). That means, \( d \) = 9. Adjacency matrix \( \mathbf{A} \in \{0,1\}^{n \times n} \).

After message passing and aggregation, a GNN layer trained on molecular Graph G can be written as follows:

\begin{equation}
    \mathbf{H}^{(k)} = \sigma \left( \mathbf{A} \mathbf{H}^{(k-1)} \mathbf{W}^{(k)} + \mathbf{b}^{(k)} \right).
\end{equation}
 \( \mathbf{W}^{(k)} \) is a learnable weight matrix, \( \mathbf{b}^{(k)} \) is a bias term, \( \sigma \) is an activation function, \( \mathbf{H}^{(k)} \in \mathbb{R}^{n \times d} \) is the matrix of hidden states for all nodes at the \(k\)-th layer, and \( \mathbf{H}^{(0)}\) is \( \mathbf{X} \).

After convolution, the generated node embeddings (i.e., atom embedding in our case) from the GNN layers are fed into the readout layer (concatenates mean and max pooling) for aggregation and later for producing graph-level representation (or drug molecular embedding): 

\begin{equation}
    \mathbf{h}_G = \text{Pooling}_{max}(\mathbf{H}^{(K)})+\text{Pooling}_{min}(\mathbf{H}^{(K)}).
\end{equation}

Another input to our hybrid network, fingerprint F, is transformed into a dense feature representation using a linear layer. The dense features afterward are combined with the output $\mathbf{h}_G$ of the readout layer using an MLP layer:

\begin{equation}
    \mathbf{F}^{*} = \mathbf{F} \mathbf{W}^{(P)} + \mathbf{c},
\end{equation}

\begin{equation}
    \mathbf{Z} = (\mathbf{h}_G + \mathbf{F}^{*}) \mathbf{W}^{(Q)} + \mathbf{d},
\end{equation}

\begin{equation}
    \mathbf{y}_{pred} = \text{softmax}(\mathbf{Z} \mathbf{W}^{R} + \mathbf{e}).
\end{equation}

$\mathbf{W}^{(P)}$, $\mathbf{W}^{Q}$, and $\mathbf{W}^{(R)}$ are weight matrices, and $\mathbf{c}$, $\mathbf{d}$, and $\mathbf{e}$ are biases of the linear layers used.

The output dimension and loss function depend on the tasks. We used MSELoss for multiregression tasks in which the model output size was 330 for all the given datasets. We ran both regression and classification for 400 epochs. We used BCELoss for multilabel classification, and the model output sizes were 3,816, 4,070, 2,801, 3,802, and 4,788 for the \textit{data1}, \textit{data2}, \textit{data3}, \textit{data4}, and \textit{data5} datasets, respectively.

\subsection{Measuring Multilabel Data Imbalance}\label{subsec:dataimbal}
\begin{figure*}
    \centering
    \rotatebox[origin=l]{90}{\makebox[0in]{\small Number of labels (ordered)}}
    \subfloat[Percentage of Samples, total (10k)]{\includegraphics[width=0.35\textwidth]{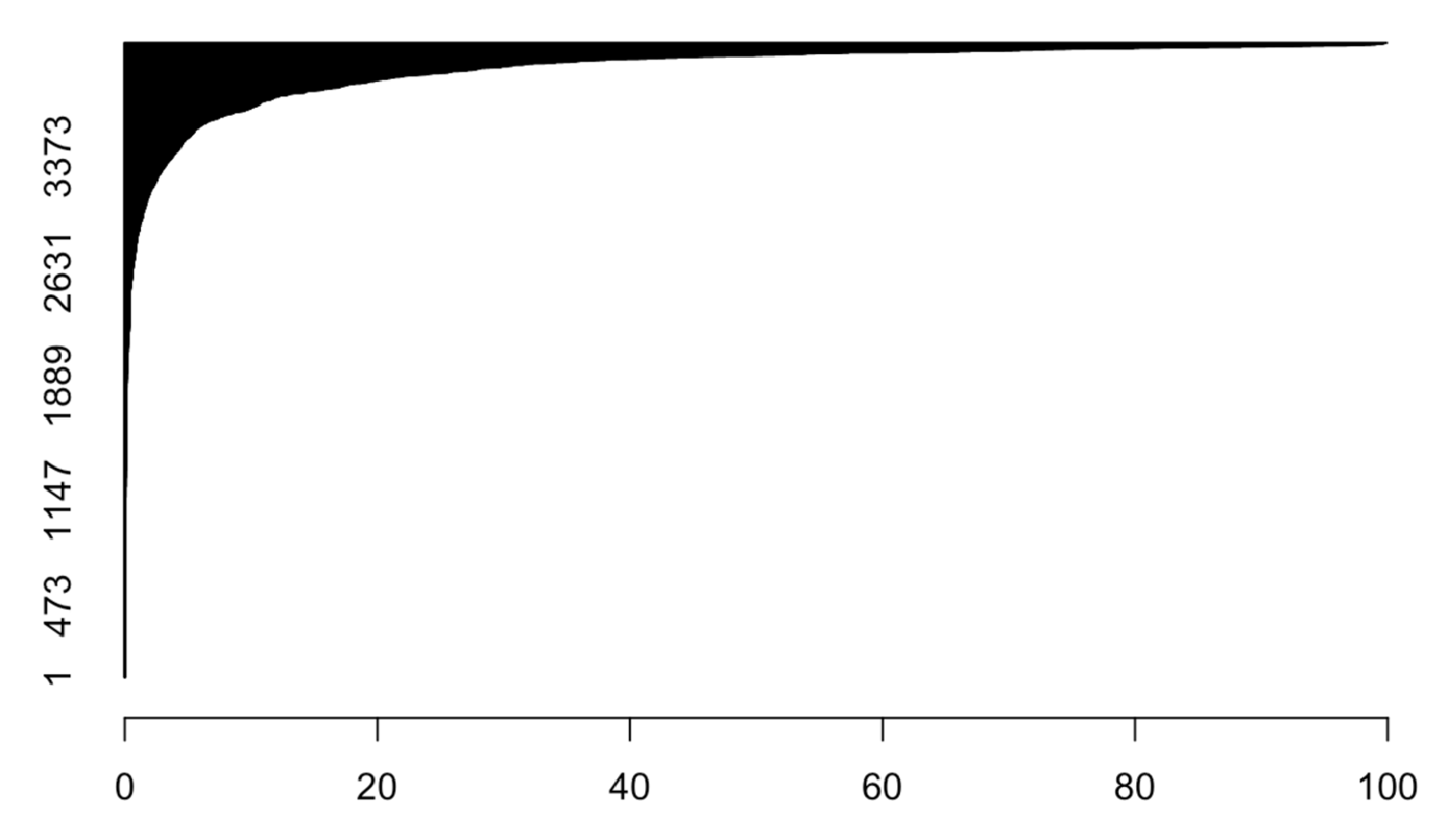}}
    \subfloat[Percentage of Samples, total (10k)]{\includegraphics[width=0.35\textwidth]{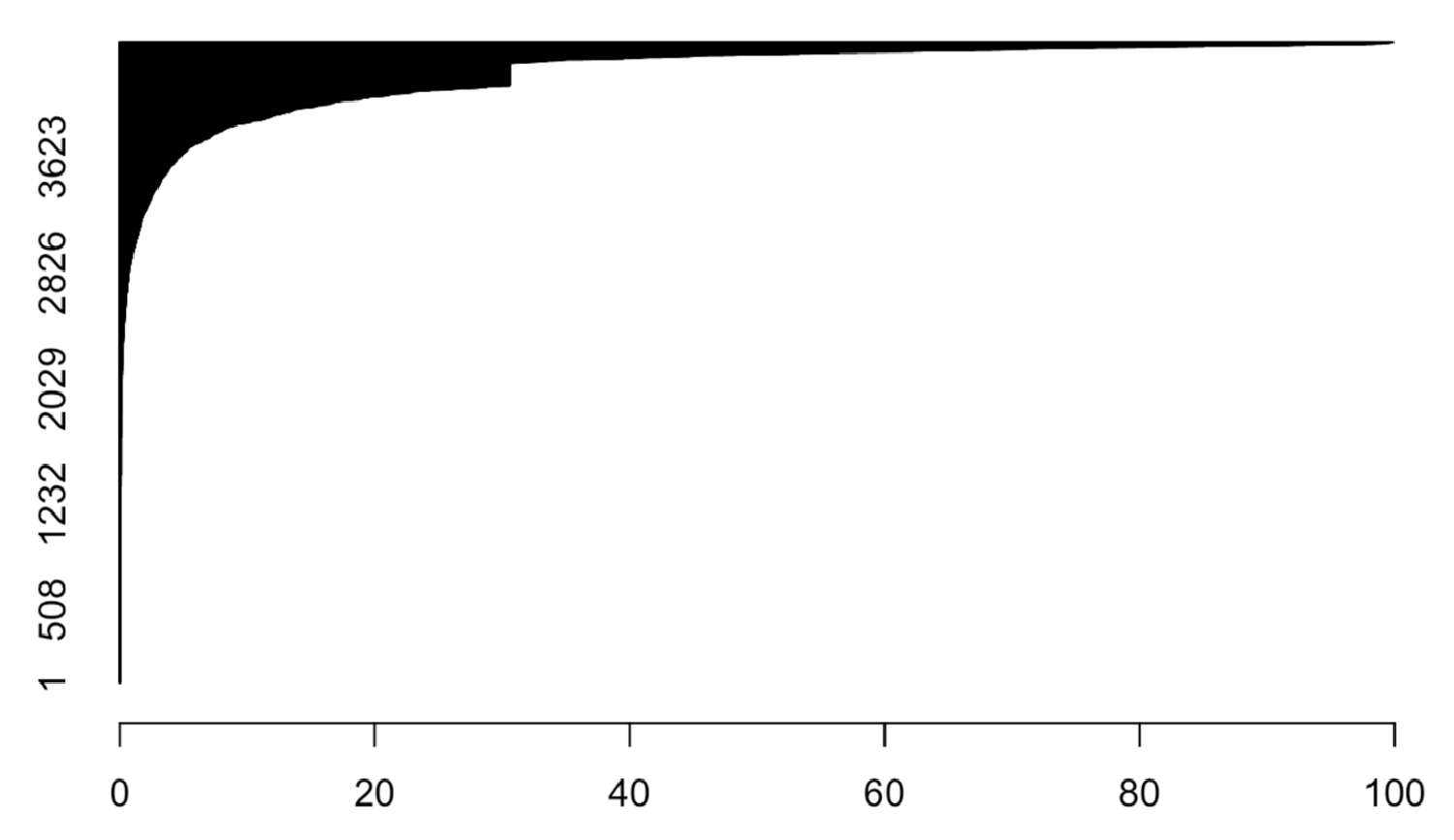}}\\
    \subfloat[Percentage of Samples, total (10k)]{\includegraphics[width=0.35\textwidth]{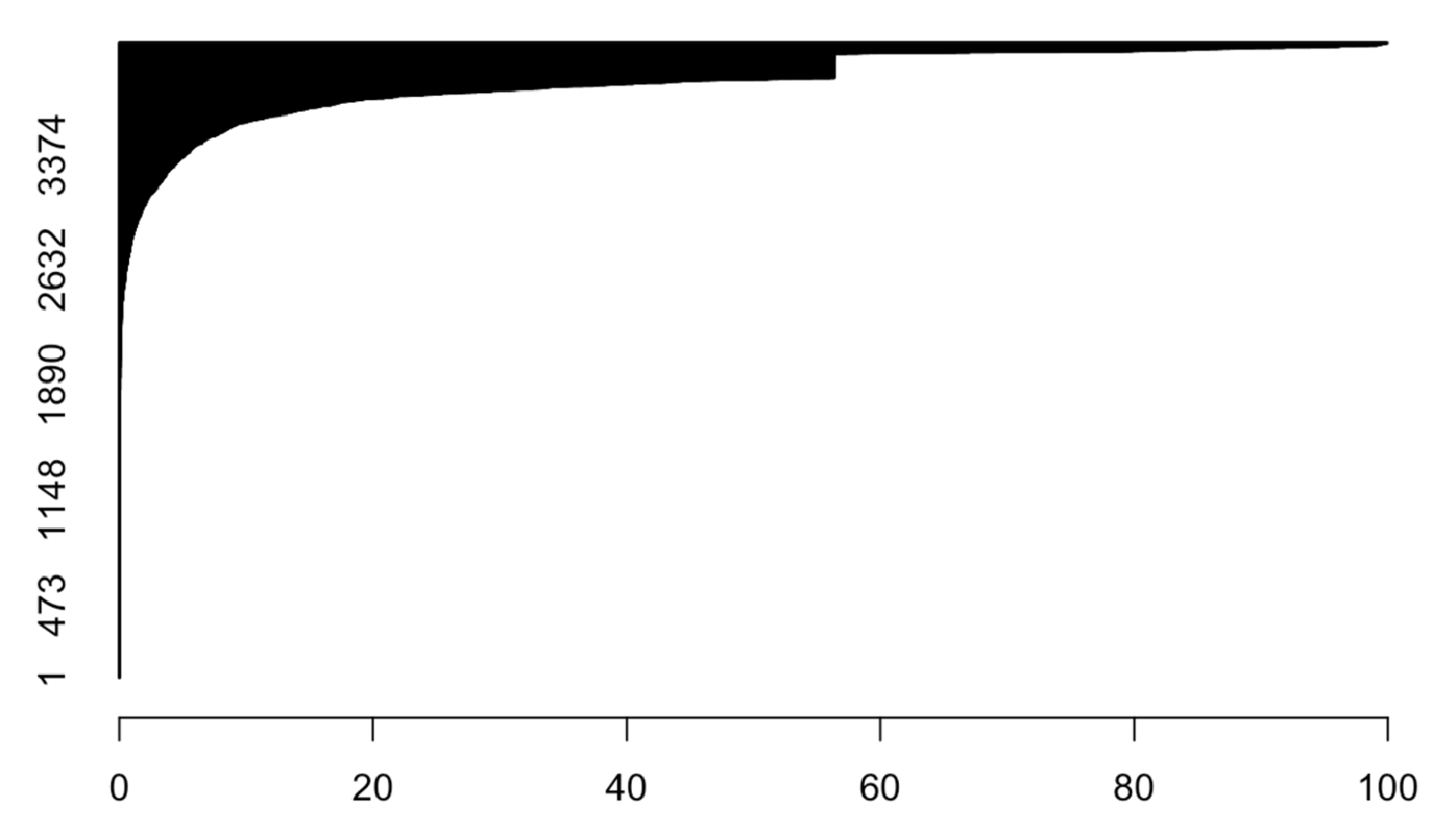}}
    \subfloat[Percentage of Samples, total (10k)]{\includegraphics[width=0.35\textwidth]{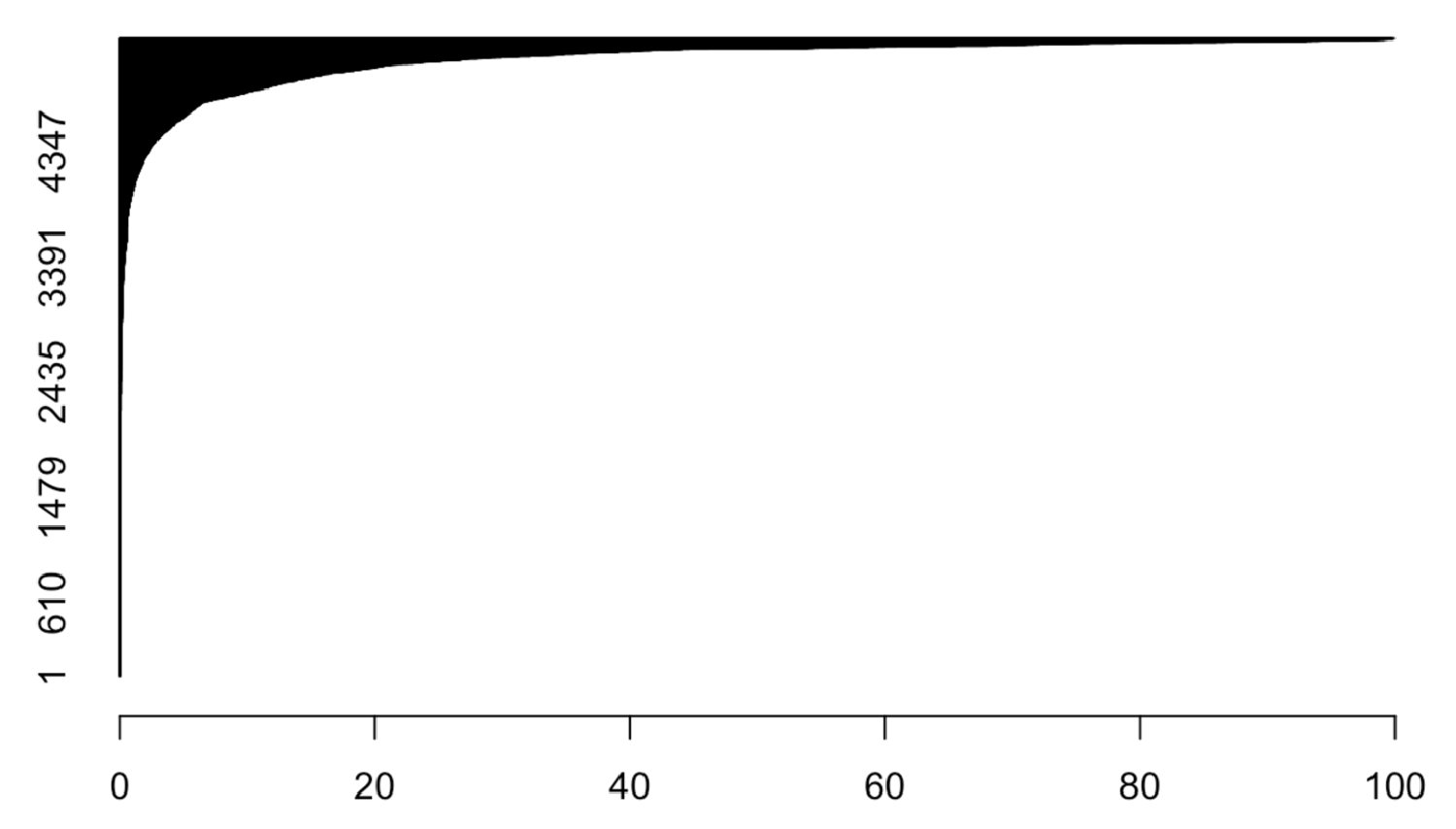}}
    \caption{Subfigures (a--d) display the percentages of data samples for each label in the datasets (\textit{data1}, \textit{data2}, \textit{data4}, and \textit{data5}).}
    \label{fig:imbal_plot}
\end{figure*}

To achieve better performance from the GNN models, addressing the issue of high data imbalance with asymmetric co-occurrence for multilabel classification takes precedence over selecting representative GNN models. Therefore, we attempt to understand the multilabel data imbalance of the datasets holistically. \textbf{Figure} \ref{fig:imbal_plot} (a--d) shows the imbalance ratio of the given datasets, where the x-axis at the bottom of each subfigure represents the percentage of samples in which a dataset label (positive effect) appears. The y-axis of each subfigure represents the range of label numbers in each dataset. All the subfigures exhibit severe high data imbalance by plotting a skewed horizontal curve from the y-axis. That means most of the labels in these datasets exist in only a few of their instances.

\subsection{Proposed Oversampling Technique} \label{subsec:heu_algo}
The main idea of the proposed multilabel oversampling technique presented in \textbf{Algorithm} \ref{alg:heurisic} is based on determining a set of minor instances ($\texttt{minor\_ins}$) and replicating the set as per the given oversampling percentage $(P\in[0,1])$ of the total number of the original dataset's instances. We adopted a strategy from Charte et al. \cite{mlsmote} for selecting minority labels that consider a label as a minority label based on a condition $IRLbl(l) > MeanIR$. Therefore, the dataset labels with a higher IRLbl than MeanIR are considered minority labels. Line 4 in \textbf{Algorithm} \ref{alg:heurisic} obtains a set of all the minority labels $minor\_labels$ in the dataset by calculating IRLbls from line 2 and MeanIR from line 3.
To determine minor instances, line 5 iterates through each instance and obtains the positive multilabel targets from each instance. A newly introduced factor ${minor\_ins\_imbalance}$ in line 6 computes a score by dividing the number of positive targets within minority labels by the total number of positive targets for each instance. It ranks the candidate minor instances based on the computed score shown in line 8 and then selects the top $(P/R)$ percent of minor instances shown in line 9. Line 10 shows how the proposed technique oversamples the minor instances ($\texttt{minor\_ins}$) by replicating each of them up to the following number $(R)$, where $R\in\mathbb{N}$ is the replication count for each minor instance in $\texttt{minor\_ins}$. Finally, line 11 produces the oversampled dataset $\hat{D}$ by adding $\texttt{new\_minor\_ins}$ to the original dataset $D$. Multiplying the top instance selection and replication count $(R)$ keeps the oversampling percentage (P) consistent as follows: $(P/R)*(R) = P$. The proposed oversampling technique has $O(n)$ time complexity, which is better than $O(n^2)$ of MLSMOTE.

In our study, we used an oversampling percentage (P) 25\% in all the oversampled datasets generated by the proposed oversampling technique. For fair comparison and analysis with the proposed oversampling technique, we generated synthetic instances of 25\% of the number of instances in each original dataset for MLSMOTE.

\renewcommand{\algorithmicrequire}{\textbf{Input:}}
\renewcommand{\algorithmicensure}{\textbf{Output:}}
\begin{algorithm}[hptb]
\scriptsize
\caption{Proposed oversampling algorithm}
\label{alg:heurisic}
\begin{algorithmic}[1]
\Require
 \Statex Multilabel graph dataset: D
 \Statex Percentage of new samples: P
 \Statex Replication count: R
\Ensure 
\Statex Oversampled dataset: $\hat{D}$
\State $L \gets$ Set of labels in D
\State $IRLbls \gets$  calculate imbalance ratio for all labels in $L$
\State $MeanIR \gets \frac{1}{|L|} \sum_{i=1}^{|L|} IRLbls(L_{i})$
\State $minor\_labels \gets$ subset of $L$ if $(IRLbls(L_{i}) > MeanIR)$
\For{\texttt{each $instance\_pos\_targets$ in $D$}}
        \State $\texttt{minor\_ins\_imbalance} \gets \frac{count(minor\_labels \: \in \: instance\_pos\_targets)}{count(instance\_pos\_targets)}$
\EndFor
\State $\texttt{minor\_ins\_imbalance} \gets rank\_desc(\texttt{minor\_ins\_imbalance})$
\State $\texttt{minor\_ins} \gets select\_top(\texttt{minor\_ins\_imbalance}, \frac{P}{R})$
\State $\texttt{new\_minor\_ins}\gets over\_sample(\texttt{minor\_ins}, (R))$
\State $\hat{D} = D$ + $\texttt{new\_minor\_ins}$
\end{algorithmic}
\end{algorithm}

\subsection{Comparing the Proposed Oversampling Technique with MLSMOTE}
To quantify the data imbalance ratio, we adopted standard imbalance metrics from Charte et al. \cite{charte2013first} and compared our proposed oversampling technique with MLSMOTE \cite{mlsmote} by using the metrics. \textbf{Table} \ref{Tab:data_oversamp} presents three dataset properties: (1) the number of labels of each given dataset; (2) the cardinality (Card), which is the average number of positive labels per sample; and (3) an average of IRLbl (Mean IRLbl), which is defined by Charte et al. \cite{mlsmote} and measures individual label imbalance ratio for each label. We report the Card and Mean IRLbl of each dataset before and after oversampling them with MLSMOTE and the proposed oversampling technique. We can observe that MLSMOTE oversampling reduces the Card (average number of labels per sample), thereby indicating information loss, whereas the proposed oversampling keeps the Card unchanged. In terms of Mean IRLbl, for the first two datasets, MLSMOTE reduces imbalance better than the proposed oversampling but performs worse for the last three datasets by a significant margin.

\begin{table}
\centering
\caption{Properties of the datasets used in the experimentation.}
\setlength{\tabcolsep}{1 pt}
\begin{tabular}{|c|c|ccc|c|ccc|}
\hline
& & \multicolumn{3}{c|}{Card} & & \multicolumn{3}{c|}{Mean IRLbl}\\
 \cline{3-5} \cline{7-9}
\multirow{2}{3.2em}{Datasets} & \multirow{2}{2.2em}{Label count} & {Before} & {Proposed} & {MLSMOTE} & & {Before} & {Proposed} & {MLSMOTE}\\
 & &  & & &  & & & \\
\noalign{\hrule}
\hline
\textit{data1} & 3,816 & 0.1176 & 0.1176 & 0.1157 & & 1,934.234 &  1,534.038 & 1,395.649\\
\textit{data2} & 4,070 & 0.1103 & 0.1103 & 0.1078 & & 2,173.156 & 1,916.821 & 1,556.54\\
\textit{data3} & 2,801 & 0.1602 & 0.1602 & 0.1578 & & 2,412.466 & 1,013.25 & 1,616.443\\
\textit{data4} & 3,802 & 0.1180 & 0.1180 & 0.1153 & & 3,216.01 & 1,386.693 & 2,112.583\\
\textit{data5} & 4,788 & 0.0937 & 0.0937 & 0.0920 & & 2,686.661 & 1,282.689 & 1,868.29\\ 
\hline
\end{tabular}
\label{Tab:data_oversamp}
\end{table}
\setlength{\textfloatsep}{3.0pt}

\subsection{Measuring Co-Occurrence of Multilabels}\label{subsec:cooccur}
As discussed in Section \ref{sec:inro}, label co-occurrence is another obstacle of a multilabel dataset and poses a significant challenge to the GNN models to train as it impacts gradient updates of model weights. This phenomenon of a multilabel dataset can be estimated by the SCUMBLE metric \cite{remedial}. The SCUMBLE metric provides a notion of co-occurrence between labels; a higher SCUMBLE indicates a higher number of instances with a joint presence of both minority and majority labels in the datasets. Therefore, this metric should be kept to a minimum for the values of the original dataset. \textbf{Table} \ref{Tab:imbal_scumble} provides the SCUMBLE values of 10 randomly selected labels from 2 datasets (\textit{data1} and \textit{data2}) for MLSMOTE and the proposed oversampling technique. MLSMOTE increases SCUMBLE values for all the randomly selected labels in the \textit{data1} dataset. A similar trend can be observed from the \textit{data2} dataset except for the first (\textit{c2515}) and last (\textit{c2614}) labels.

\begin{table}
\centering
\caption{Mean SCUMBLE values of oversampling methods on 10 random labels for 2 datasets.}
\setlength{\tabcolsep}{1.25 pt}
\begin{tabular}{|c|c|c|c|c|c|c|c|c|}
\hline
& \multicolumn{3}{c|}{\textit{data1} dataset} & & & \multicolumn{3}{c|}{\textit{data2} dataset}\\
 \cline{1-4} \cline{6-9}
{Labels} & {Original} & {Proposed} & {MLSMOTE} & & {Labels} & {Original} & {Proposed} & {MLSMOTE}\\
\noalign{\hrule}
\hline
\textit{c1057} & 0.2057 & 0.2056 & 0.2420 & & \textit{c2515} & 0.8241 &  0.8321 & 0.8180\\
\textit{c1733} & 0.6487 & 0.3763 & 0.5342 & & \textit{c3508} & 0.3422 & 0.2983 & 0.3786\\
\textit{c1550} & 0.1151 & 0.1084 & 0.1103 & & \textit{c1428} & 0.0672 & 0.0558 & 0.0655\\
\textit{c1276} & 0.1615 & 0.1097 & 0.1825 & & \textit{c271} & 0.0786 & 0.0732 & 0.0789\\
\textit{c1516} & 0.9451 & 0.8769 & 0.9323 & & \textit{c1467} & 0.6297 & 0.5999 & 0.6569\\
\textit{c3805} & 0.7545 & 0.6388 & 0.7212 & & \textit{c2577} & 0.2564 &  0.1558 & 0.2534\\
\textit{c3375} & 0.1591 & 0.1330 & 0.1867 & & \textit{c2067} & 0.0000 & 0.0000 & 0.0000\\
\textit{c2127} & 0.4808 & 0.4871 & 0.4987 & & \textit{c3037} & 0.4915 & 0.3943 & 0.3967\\
\textit{c1741} & 0.4881 & 0.4758 & 0.5033 & & \textit{c1225} & 0.7449 & 0.7538 & 0.7552\\
\textit{c3697} & 0.9767 & 0.9458 & 0.9718 & & \textit{c2614} & 0.0967 & 0.0970 & 0.0920\\ 
\hline
\end{tabular}
\label{Tab:imbal_scumble}
\end{table}
\setlength{\textfloatsep}{3.0pt}

\begin{figure*}
    \centering
    \subfloat[10 random labels from \textit{data1} original data]{\includegraphics[width=0.33\textwidth]{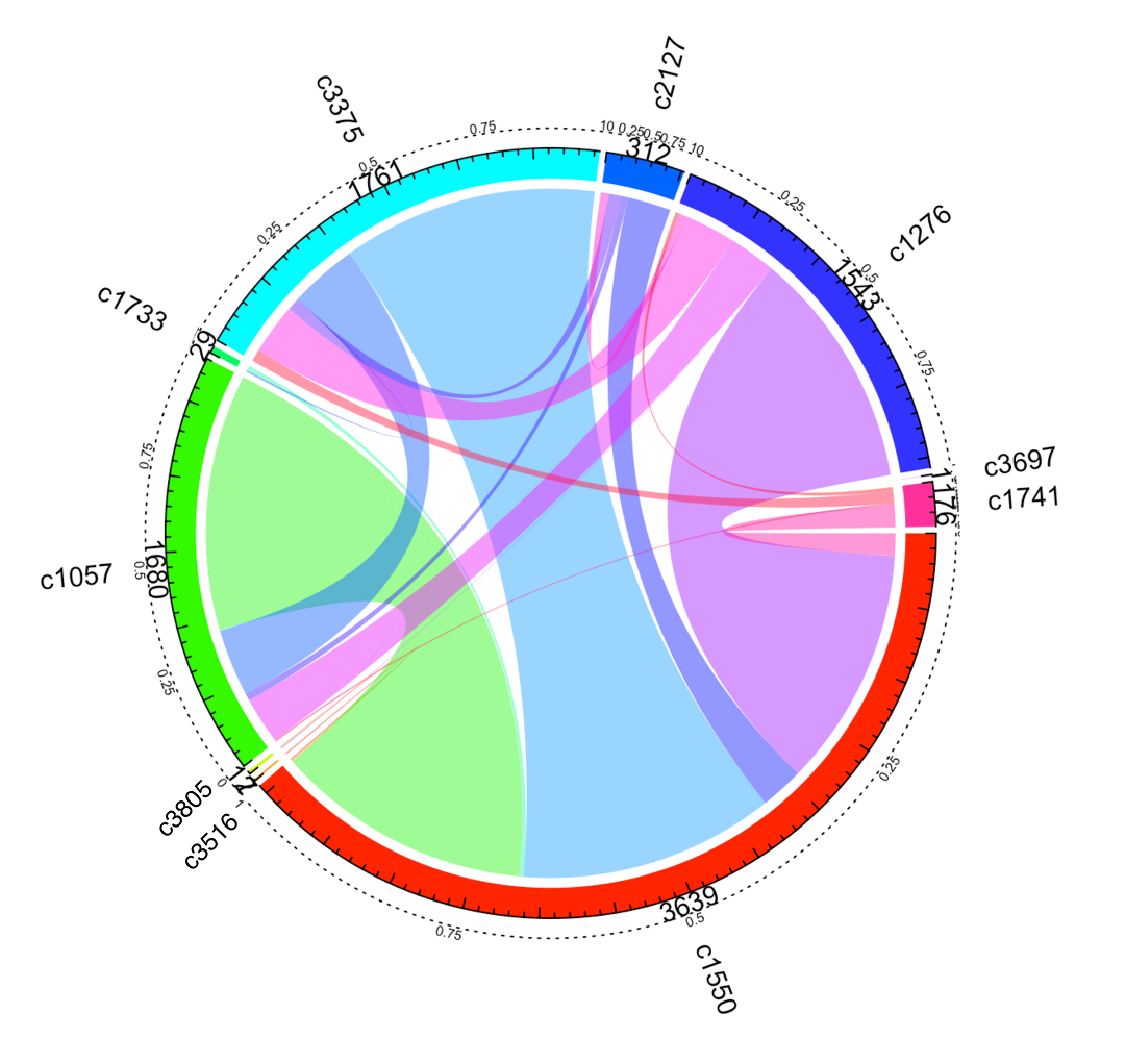}}
    \subfloat[Same 10 labels of \textit{data1}data with the \\ .\hspace{1cm} proposed oversampling]{\includegraphics[width=0.33\textwidth]{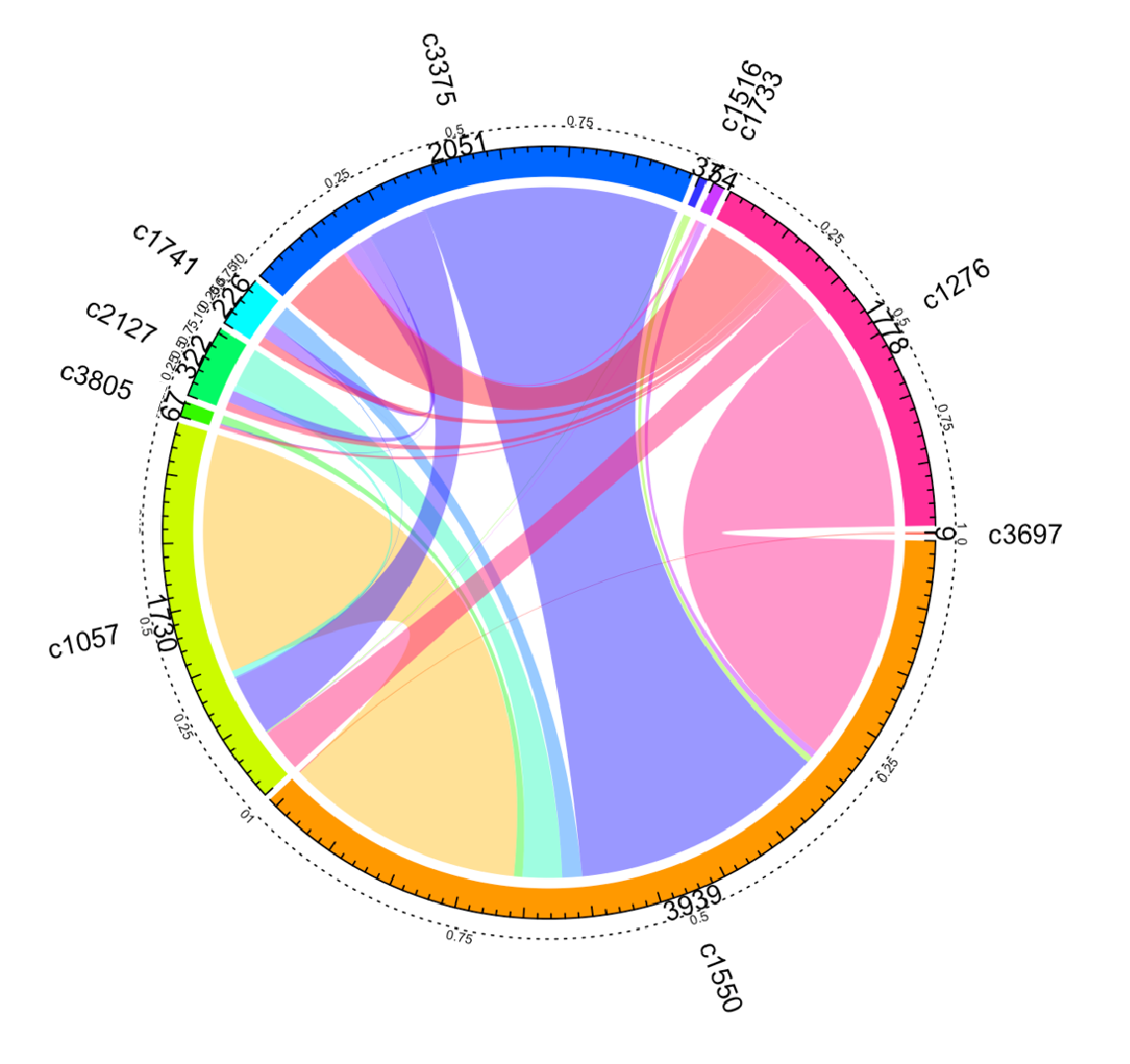}}
    \subfloat[10 random labels of \textit{data1} data with \\ .\hspace{1cm} MLSMOTE oversampling]{\includegraphics[width=0.33\textwidth]{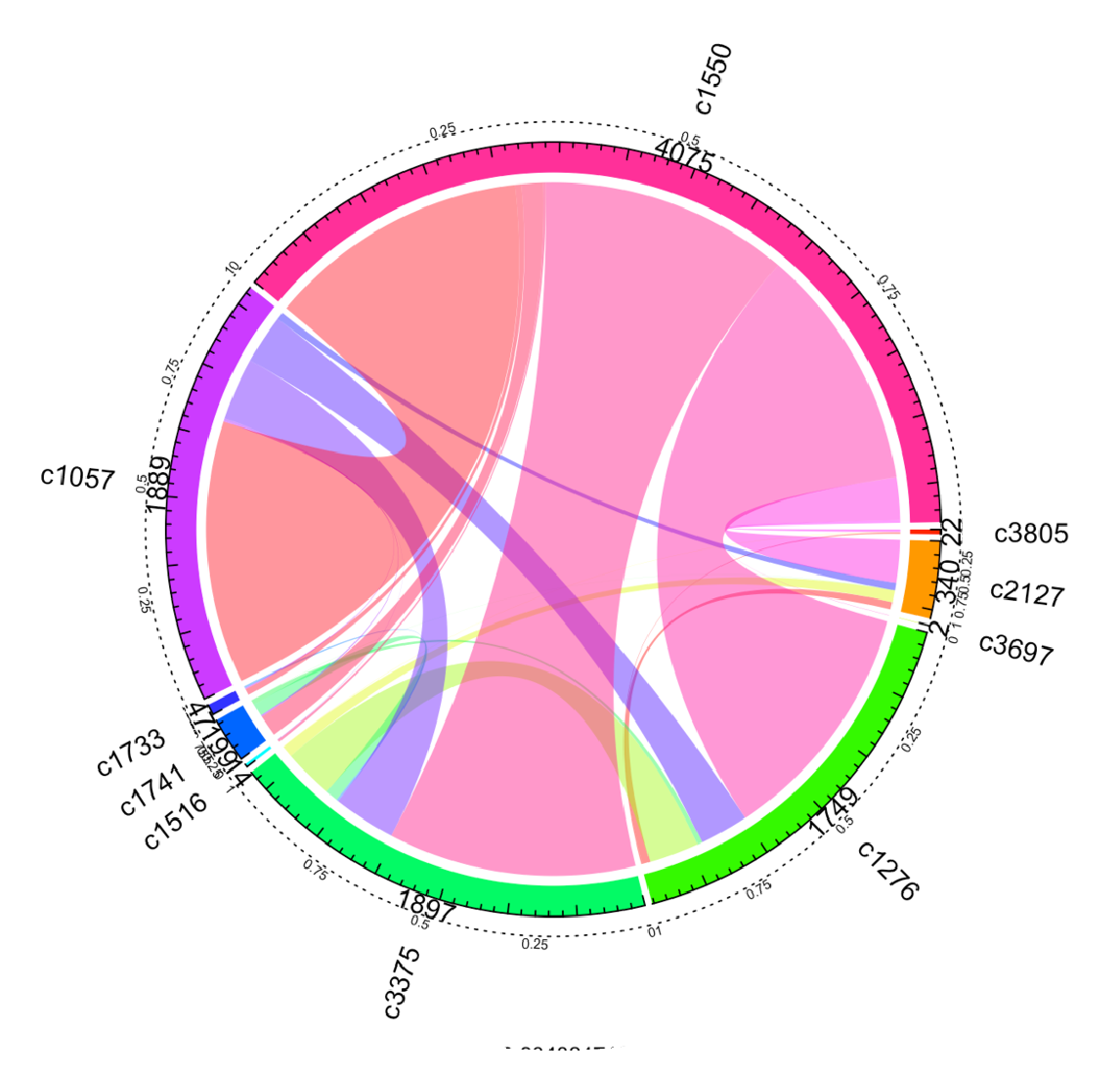}} \\
    \subfloat[10 random labels from \textit{data2} original data]{\includegraphics[width=0.33\textwidth]{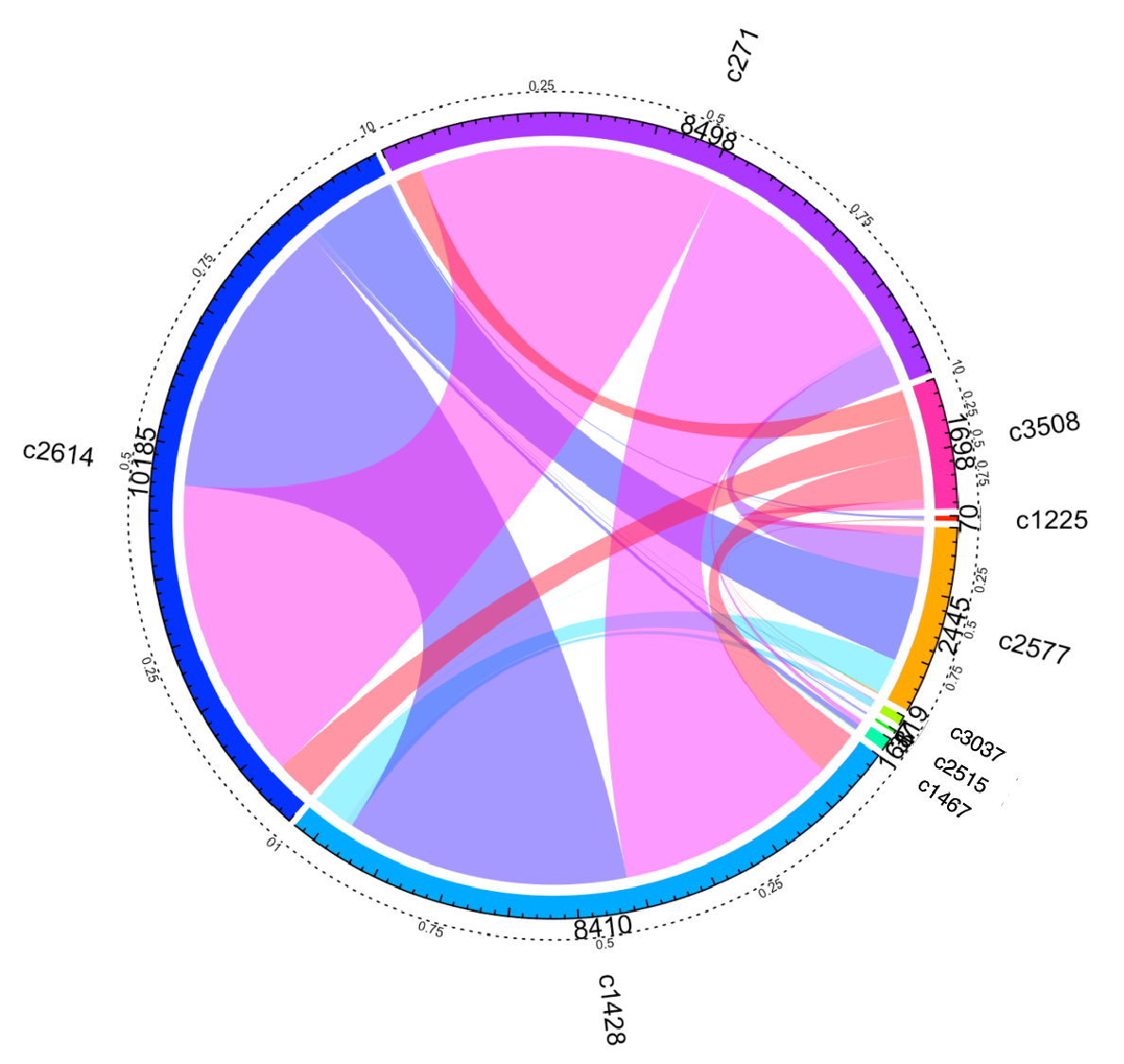}}
    \subfloat[Same 10 labels of \textit{data2} data with the \\ .\hspace{1cm} proposed oversampling]{\includegraphics[width=0.33\textwidth]{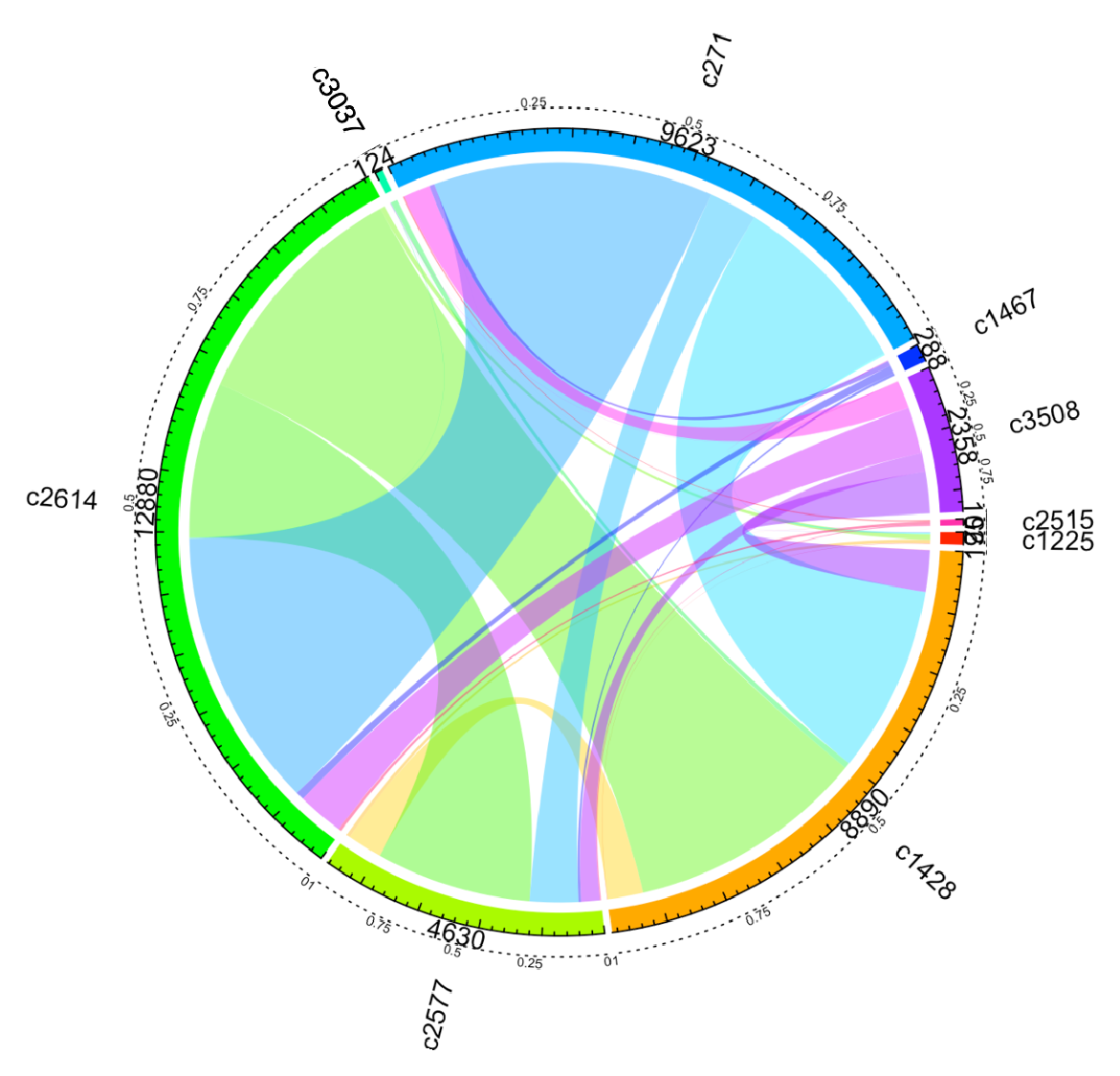}}
    \subfloat[10 random labels of \textit{data2} data with \\ .\hspace{1cm} MLSMOTE oversampling]{\includegraphics[width=0.33\textwidth]{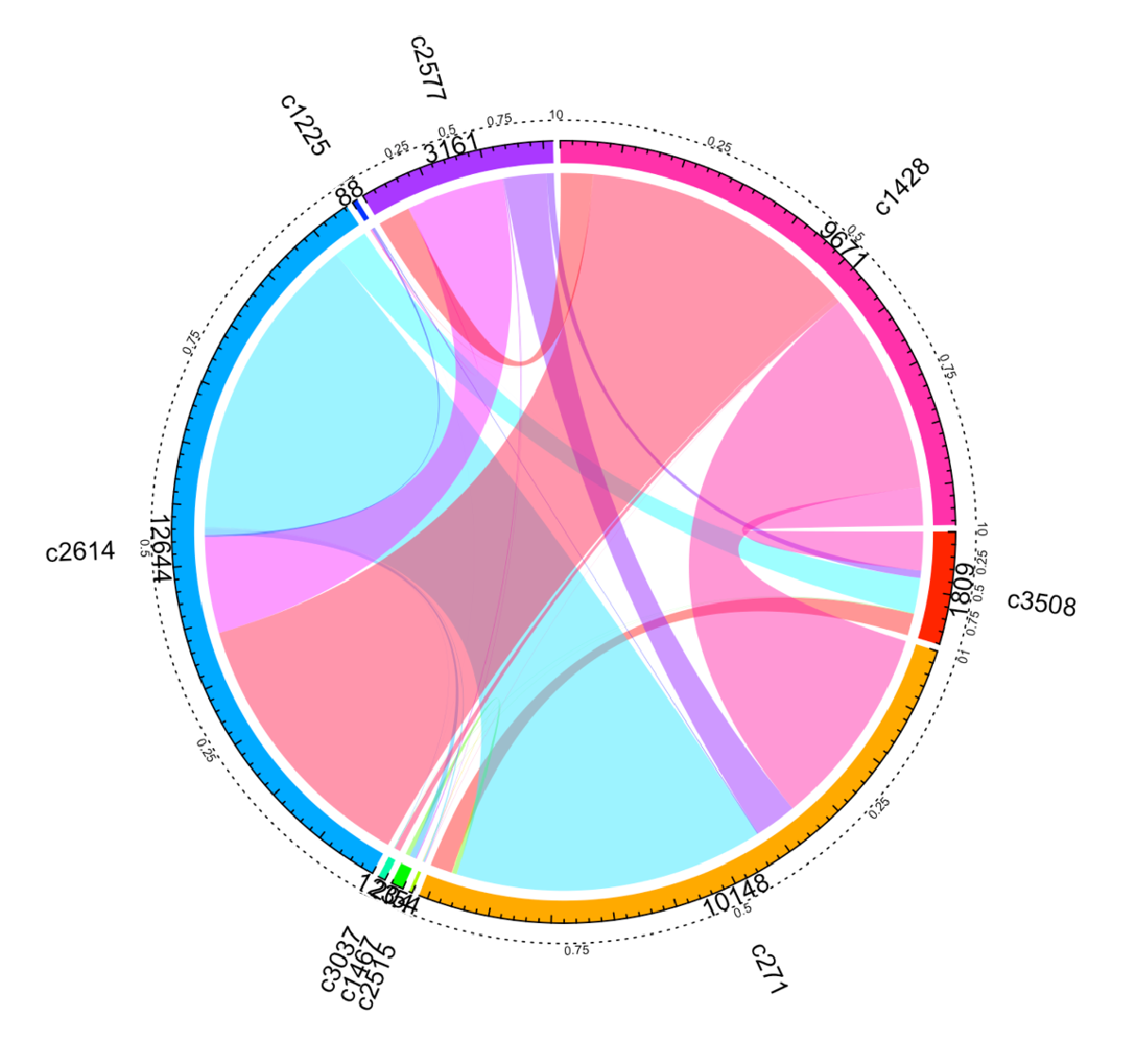}}
    \caption{Subfigures (a--f) display co-occurrence of labels for two oversampling methods in datasets \textit{data1} and \textit{data2}.}
    \label{fig:cooccur_plot}
\end{figure*}

To visualize the co-occurrence among the minority and majority labels, we present co-occurrence plots in \ref{fig:cooccur_plot} by using the same labels presented in \textbf{Table} \ref{Tab:imbal_scumble} for the \textit{data1} dataset and \textit{data2} dataset. The uniquely color-coded arcs in each subplot correspond to individual labels present in the multilabel sets of the given dataset. The size of each arc indicates the number of instances in which a label appears in respective datasets. The color-coded links represent the interactions among labels in which the thickness of the link connecting one arc to another is proportional to the number of joint occurrences within the dataset instances.
\textbf{Figure} \ref{fig:cooccur_plot}(a) indicates that minority label (\textit{c1741}) co-occurs with label \textit{c1550} in more than half of the total (total 176 instances) number of instances, which is the most frequent major label (total of 3,639 instances) in this label set. After applying the proposed oversampling algorithm, we can observe the impact on the oversampled label sets from \textbf{Figure} \ref{fig:cooccur_plot}(b), in which the frequency of the minority label \textit{c1741} increases (total of 226 instances), but the co-occurrence with majority label \textit{c1550} decreases. By contrast, \textbf{Figure} \ref{fig:cooccur_plot}(c) corresponds to MLSMOTE oversampling, depicting an opposite scenario in which the algorithm could not reduce co-occurrence. That means even though the number of instances in which label \textit{c1550} count appears to increase (total of 4,075 instances for \textit{c1550}  and 199 for \textit{c1741}) for MLSMOTE, is still higher than that of the proposed oversampling (total of 3,939 instances for \textit{c1550}).

In a different context, the co-occurrence of two relatively minority labels demonstrates that the proposed oversampling technique contributes to increasing interactions among minority labels. For example, two minority labels (\textit{c2577} and \textit{c3508}) for dataset \textit{data2} show thin links between them in \textbf{Figure} \ref{fig:cooccur_plot}(d) and \textbf{Figure} \ref{fig:cooccur_plot}(f), which correspond to the original label set and MLSMOTE generated label set, respectively. However, the link between the labels can be clearly identified in \textbf{Figure} \ref{fig:cooccur_plot}(e), and the link is thicker than others, supporting the notion of having lower SCUMBLE values for labels \textit{c2577} and \textit{c3508} reported in \textbf{Table} \ref{Tab:imbal_scumble}.

\setlength{\textfloatsep}{2.0pt}

\section{Results and Discussion}
The left side of \textbf{Table} \ref{Tab:multiRegClass} presents the evaluation results of the multiregression tasks for four standard GNN models (GCN, GAT, GMM, and AttentiveFP) with their hybrid variants that combine graph inputs with respective molecular fingerprints (atom-pair fingerprint). We report the mean absolute error (MAE) and Pearson correlation $R^2$ between predicted and target values for each GNN model. We selected the atom-pair fingerprint \cite{atompair_finger} for the hybrid models because it performs better than MACCS-keys \cite{maccs}, Avalon \cite{avalon}, and Morgan fingerprints \cite{morgan}. A high correlation $R^2$ implies that the model predictions follow the actual target values, whereas the MAE directly indicates the average absolute prediction error. Correlation does not explain average errors between predictions and actual values and the MAE depends on the units of the target variable. That means lower values of target variables cause a low MAE. However, the two metrics together provide a more comprehensive understanding of model performance, indicating the accuracy and trend of the predictions.

We observe that either the hybrid AttentiveFP or standard AttentiveFP model has the lowest MAE in almost all the datasets except \textit{data3}. Additionally, the standard GMM model shows the highest correlation value in two datasets: \textit{data1} and \textit{data5}. The hybrid GAT and standard GAT  demonstrate the highest correlation in the \textit{data2} and \textit{data4} datasets. In general, the evaluation of the GNN models in the regression task indicates that hybrid AttentiveFP performs better overall than other models for this dataset. Notably, Table \ref{Tab:multiRegClass} indicates that combining fingerprint input into the GNN models significantly improves the MAE for all datasets.

\begin{table*}
\centering
\caption{Evaluation of GNN models in multiregression and multilabel classification tasks on drug effect datasets.}
\setlength{\tabcolsep}{5 pt}
\begin{tabular}{|c|c|cc|cc|c|ccc|ccc|}
\hline
& & \multicolumn{4}{|c|}{Multi Regression task} & & \multicolumn{6}{c|}{Multilabel Classification task}\\
 \cline{3-6} \cline{8-13}
 
& & \multicolumn{2}{|c|}{Only graph input} & \multicolumn{2}{c|}{(Graph+fingerprint) input} & & \multicolumn{3}{c|}{Only graph input} & \multicolumn{3}{c|}{(Graph+fingerprint) input} \\
 \cline{3-4} \cline{5-6} \cline{8-10} \cline{11-13}
 
{Datasets} & {Models} & {Correlation} & {MAE} & {Correlation} & {MAE} & & Precision  & Recall & F1 & Precision & Recall & F1 \\
\noalign{\hrule}
\hline
\multirow{3}{6em}{\textit{data1}} & GCN & 0.9410 & 0.3087 & 0.9550 & 0.0271 & & 0.0912 & 0.1085 & 0.0981 & 0.1383 & 0.1287 & 0.1270\\
 & GAT & 0.9786 & 0.2843 & 0.9394 & 0.0138  & & 0.0911 & 0.1077 & 0.0978 & 0.1384 & 0.1284 & 0.1270\\
 & GMM & \textbf{0.9790} & 0.1599 & 0.9718 & 0.0166  & & 0.0945 & 0.1059 & 0.0985  & 0.1384 & 0.1282 & 0.1268\\
 & AttentiveFP & 0.9165 & 0.0265 & 0.9422 & \textbf{0.0114}  & & 0.1400 & 0.1230 & 0.1227 & \textbf{0.1535} & \textbf{0.1292} & \textbf{0.1306}\\
\hline
\multirow{3}{6em}{\textit{data2}} & GCN & 0.8063 & 0.4257 & 0.9573 & 0.0155 & & 0.0744 & 0.0949 & 0.0819 & 0.1569 & 0.1434 & 0.1427\\
 & GAT & 0.9636 & 0.3984 & \textbf{0.9914} & 0.0192 & & 0.0735 & 0.0946 & 0.0815  & 0.1571 & 0.1430 & 0.1424 \\
 & GMM & 0.9394 & 0.1275 & 0.8253 & 0.0759 & & 0.1052 & 0.1021 & 0.0957  & 0.1562 & 0.1440 & 0.1434\\
 & AttentiveFP & 0.9827 & 0.0140 & 0.9147 & \textbf{0.0122} & & 0.1653 & 0.1433 & 0.1450 &  \textbf{0.1707} & \textbf{0.1469} & \textbf{0.1498}\\
\hline
\multirow{3}{6em}{\textit{data3}} & GCN & 0.4953 & 0.1714 & 0.4750 & 0.0761 & & 0.1332 & 0.1481 & 0.1389 & 0.1876 & 0.1733 & 0.1731\\
 & GAT & 0.4015 & 0.1993 & 0.5473 & 0.0732 & & 0.1318 & 0.1444 & 0.1366 & 0.1869 & 0.1719 & 0.1715 \\
 & GMM & 0.5236 & 0.2247 & 0.5816 & \textbf{0.0667} & & 0.1383 & 0.1494 & 0.1409 & 0.1881 & \textbf{0.1736} & \textbf{0.1733}\\
 & AttentiveFP & \textbf{0.5287} & 0.0728 & 0.4802 & 0.0722 & & 0.1855 & 0.1652 & 0.1673 & \textbf{0.1943} & 0.1702 & 0.1732\\
 \hline
\multirow{3}{6em}{\textit{data4}} & GCN & 0.7208 & 0.2214 & 0.6955 & 0.1015 & & 0.0796 & 0.1059 & 0.0889 & 0.1615 & 0.1449 & 0.1456 \\
 & GAT & \textbf{0.8299} & 0.2011 & 0.7940 & 0.0540 & & 0.0774 & 0.1005 & 0.0858 &  0.1620 & 0.1447 & 0.1457\\
 & GMM & 0.7192 & 0.2100 & 0.8036 & 0.0676 & & 0.0871 & 0.1057 & 0.0910 & 0.1627 & 0.1459 & 0.1468\\
 & AttentiveFP & 0.6477 & \textbf{0.0346} & 0.7709 & 0.0364 & & 0.1713 & \textbf{0.1474} & 0.1492 & \textbf{0.1773} & 0.1468 & \textbf{0.1505}\\
\hline
\multirow{3}{6em}{\textit{data5}} & GCN & 0.5707 & 0.2879 & 0.7857 & 0.0816 & & 0.0708 & 0.0827 & 0.0755 & 0.1124 & 0.1022 & 0.1013\\
 & GAT & 0.7920 & 0.2047 & 0.6570 & 0.0750 & & 0.0715 & 0.0843 & 0.0767 & 0.1123 & 0.1029 & 0.1018\\
 & GMM & \textbf{0.8045} & 0.2678 & 0.7338 & 0.0805 & & 0.0740 & 0.0854 & 0.0777 & 0.1141 & \textbf{0.1041} & 0.1033\\
 & AttentiveFP & 0.6933 & 0.0688 & 0.7546 & \textbf{0.0448} & & 0.1134 & 0.1003 & 0.1005 & \textbf{0.1205} & 0.1031 & \textbf{0.1042}\\
\hline
\end{tabular}
\label{Tab:multiRegClass}
\end{table*}
\setlength{\textfloatsep}{2.0pt}

\begin{figure*}
    \centering
    \subfloat[Scatter plot for \textit{data1}]{\includegraphics[width=0.35\textwidth]{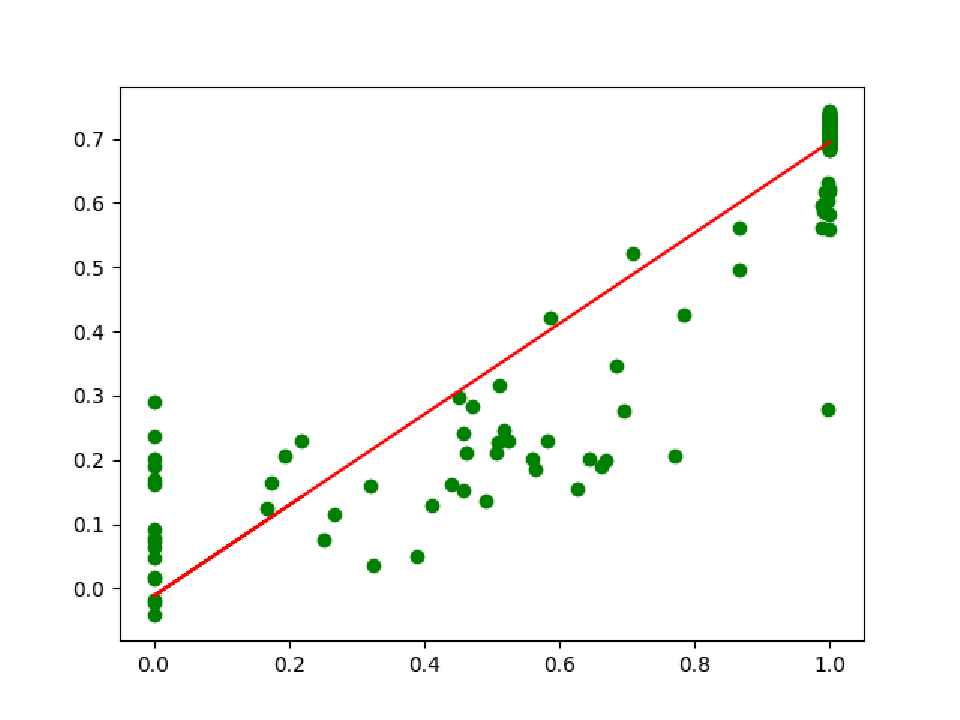}}
    \hspace*{-2em}
    \subfloat[Scatter plot for \textit{data2}]{\includegraphics[width=0.35\textwidth]{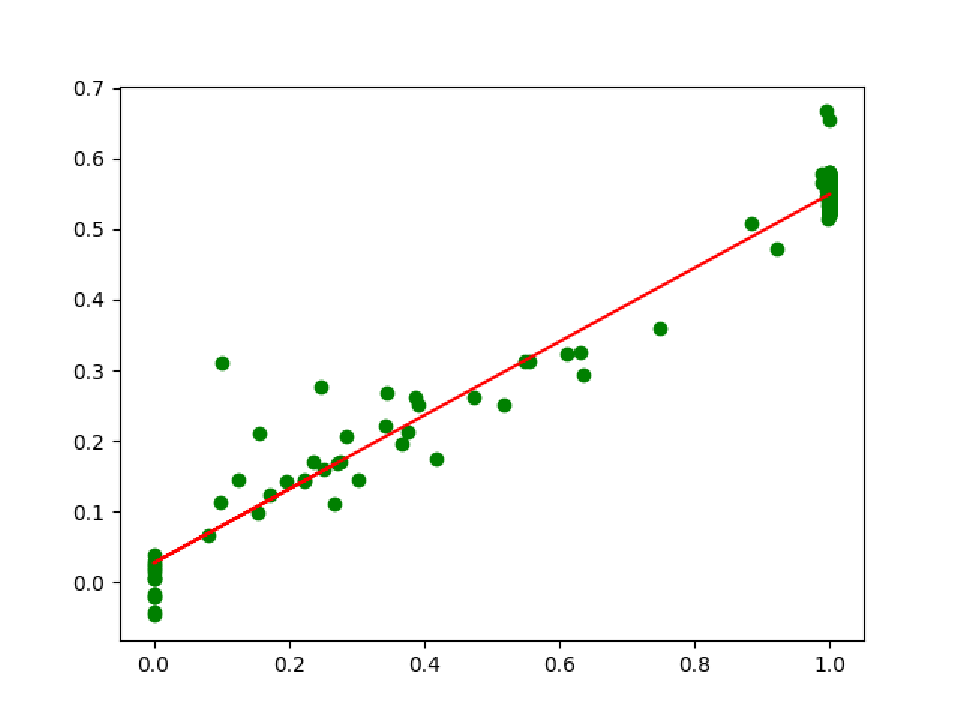}} .
    \subfloat[Scatter plot for \textit{data4}]{\includegraphics[width=0.35\textwidth]{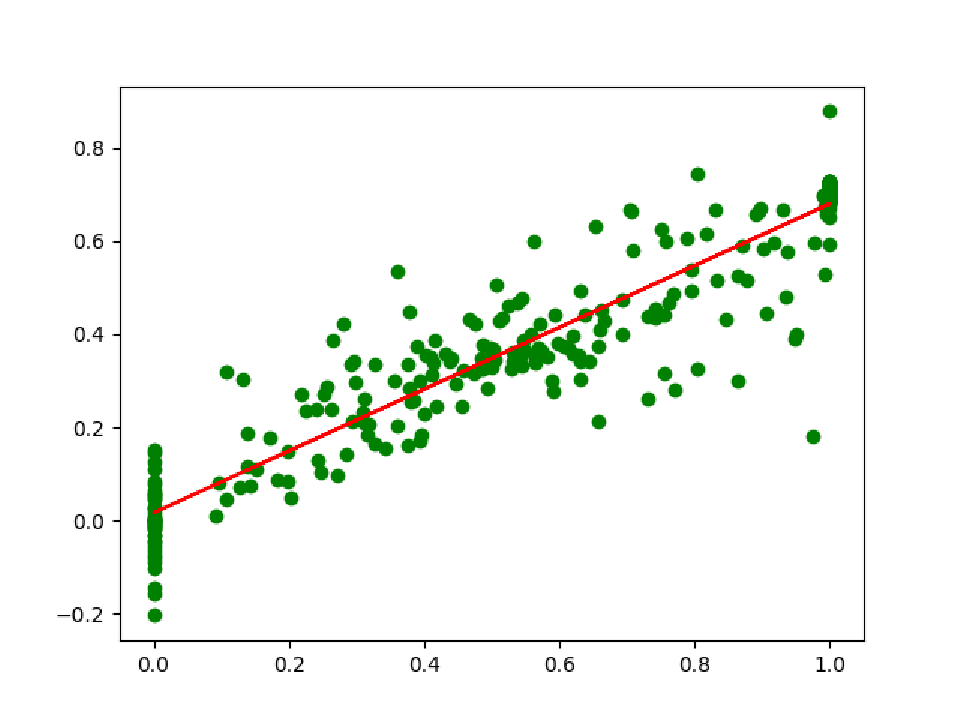}}
    \hspace*{-2em}
    \subfloat[Scatter plot for \textit{data5}]{\includegraphics[width=0.35\textwidth]{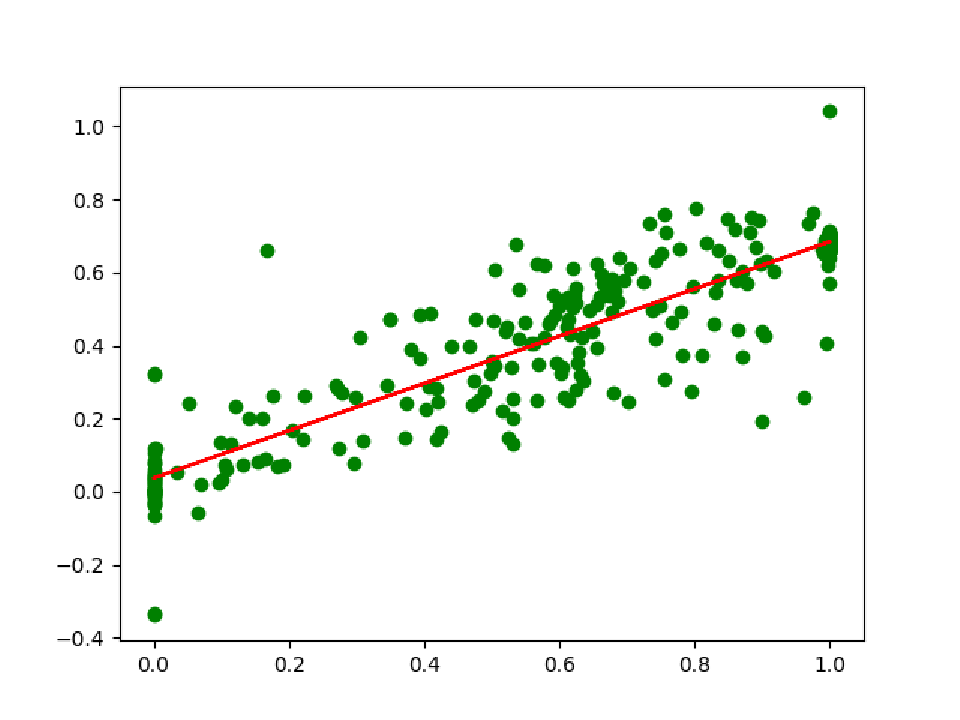}}
    \caption{Scatter plots (a--d) display the correlation between the target and predicted values for the given datasets.}
    \label{fig:corr_plot}
\end{figure*}

\textbf{Figure} \ref{fig:corr_plot}(a--d) displays scatter plots between targets and predictions for the datasets \textit{data1}, \textit{data2}, \textit{data4}, and \textit{data5} with AttentiveFP models in the multiregression task. We can observe from the plots that the predictions follow the direction of the targets.

\setlength{\textfloatsep}{2.0pt}

The right side of \textbf{Table} \ref{Tab:multiRegClass} presents evaluation measures of the multilabel classification tasks for all four standard GNN models (GCN, GAT, GMM, and AttentiveFP) with their hybrid variants for all the given datasets. All the models are severely affected by the substantial presence of high data imbalance in all these datasets. The hybrid variants of AttentiveFP perform the best on almost all the datasets in all three evaluation measures---precision, recall, and F1---although some minor discrepancies exist. Additionally, hybrid GMM performs the best on dataset \textit{data3}. In the regression task, a similar trend is observed from the F1 measures: The hybrid GNN models benefit from combining molecular fingerprints with graph inputs and perform better in multilabel classification tasks.

We leveraged the results reported in \textbf{Table}  \ref{Tab:multiRegClass}, selected hybrid AttentiveFP as the best-performing model architecture for multilabel classification, and used this as the representative architecture. Unlike traditional imbalanced datasets, a multilabel dataset poses two inherent challenges: (1) a large number of minority (majority) labels for each data instance and (2) the existence of minority and majority labels in the same instance. In our analysis, the given datasets have a much larger number of labels and higher asymmetric label co-occurrence than most other multilabel benchmark datasets. This feature of a multilabel dataset described in Section \ref{subsec:cooccur} makes graph learning far more challenging for the studied GNN models. To address the high-class imbalance and asymmetric label co-occurrence, we applied our proposed multilabel oversampling technique to all the graph datasets and then used the MLSMOTE technique \cite{mlsmote} for comparison and evaluation. We experimented with four sampling use cases: (1) fingerprint input with no resampling, (2) graph input with no resampling, (3) fingerprint input with MLSMOTE oversampling, and (4) graph input plus fingerprints with the proposed oversampling for each given dataset to perform multilabel classification and then report the result in \textbf{Table} \ref{tab:final_table}.

All the hybrid AttentiveFP models trained on the datasets oversampled by the proposed technique outperformed models trained on MLSMOTE's oversampled datasets. 

\begin{table}[hbtp]
\caption{Evaluation of multilabel classification performance on different sampled datasets. We selected (AttentiveFP + fingerprint) as our proposed hybrid GNN architecture and trained on the oversampled datasets by using our oversampling technique.}
\begin{center}
\setlength{\tabcolsep}{2 pt}
\begin{tabular}{|c|c|c|c|c|c|}
\hline
\textbf{Datasets}&\textbf{Methods} & \textbf{Precision} & \textbf{Recall} & \textbf{F1}\\
\hline
\multirow{4}{6em}{\textit{data1}} & Fingerprint input, no resampling  &  0.1055 & 0.1088 & 0.1028 \\
& Graph input, no resampling   & 0.1797 & 0.1466 & 0.1515\\
& Fingerprint input, MLSMOTE &  0.1028 & 0.1088 & 0.1031 \\
& Proposed & \textbf{0.2860} & \textbf{0.2227} & \textbf{0.2363} \\
\hline
\multirow{4}{6em}{\textit{data2}} & Fingerprint input, no resampling  &  0.1298 & 0.1263 & 0.1242 \\
& Graph input, no resampling  & 0.1858 & 0.1554 & 0.1596\\
& Fingerprint input, MLSMOTE &  0.1275 & 0.1226 & 0.1215 \\
& Proposed & \textbf{0.2451} & \textbf{0.2031} & \textbf{0.2103}\\
\hline
\multirow{4}{6em}{\textit{data3}} & Fingerprint input, no resampling & 0.1454 & 0.1468 & 0.1413 \\
& Graph input, no resampling  & 0.2318 & 0.1979 & 0.2040 \\
& Fingerprint input, MLSMOTE &  0.1433 & 0.1461 & 0.1406 \\
& Proposed & \textbf{0.3104} & \textbf{0.2383} & \textbf{0.2521} \\
\hline
\multirow{4}{6em}{\textit{data4}} & Fingerprint input, no resampling  &  0.1261 & 0.1201 & 0.1175 \\
& Graph input, no resampling  & 0.1955 & 0.1622 & 0.1665 \\
& Fingerprint input, MLSMOTE & 0.1236 & 0.1216 & 0.1191 \\
& Proposed & \textbf{0.2554} & \textbf{0.2113} & \textbf{0.2182} \\
\hline
\multirow{4}{6em}{\textit{data5}} & Fingerprint input, no resampling &  0.0811 & 0.0863 & 0.0806\\
& Graph input, no resampling  & 0.1667 & 0.1220 & 0.1300 \\
& Fingerprint input, MLSMOTE &  0.0812 & 0.0871 & 0.0822 \\
& Proposed & \textbf{0.1860} & \textbf{0.1503} & \textbf{0.1556} \\
\hline
\end{tabular}
\label{tab:final_table}
\end{center}
\end{table}
\setlength{\textfloatsep}{1.0pt}

The results presented in \textbf{Table} \ref{tab:final_table} for multilabel classification are explained as follows. The MLSMOTE algorithm performs a nearest-neighbor search for each data instance containing a minority label to interpolate a new instance. MLSMOTE algorithm has two issues. First, the molecular graph datasets do not follow the notion that instance proximity corresponds to similarity. Second, the algorithm selects all the instances that contain minority labels (many of those are not proper candidates). Therefore, oversampling via neighborhood-based interpolation for synthetic data generation and sampling from a large number of data instances is inappropriate because it can interpolate a molecule encoding with a different set of effects. Moreover, the drug molecules are structurally different, and a small difference in molecular structure results in completely different effect vectors. We observed that in almost all the experiments, either multiregression or multilabel classification from \textbf{Table} \ref{Tab:multiRegClass} indicates that combining fingerprint input into the GNN models significantly improves the MAE and F1 score for all datasets. This prediction performance improvement indicates that rich input information helps address the issues of multilabel datasets and validates the hypothesis yet unexplored but hinted by a previously published work \cite{cheminet}.



\section{Conclusion}
In this work, we studied the expressiveness of GNN models trained on drug molecular graphs. We investigated ways to improve a model’s capacity for multilabel drug effect prediction, in which the drug effects include continuous and categorical values. We observed that GNN models trained on continuous values for multiregression tasks perform well because the continuous values are almost normally distributed. By contrast, GNN models trained on categorical values for multilabel classification tasks struggle to learn from high-dimensional and asymmetrically co-occurrent multilabels. We demonstrated that replicating instances with minor labels is more effective in addressing multilabel imbalance than generating synthetic instances. Finally, we proposed a simple yet effective oversampling technique to address the existing multilabel classification issues, thereby enabling performance improvement. We plan to extend our research further to build custom GNN layers and networks to tackle high multilabel imbalance in the domain of drug molecular properties and effect prediction.

\bibliographystyle{IEEEtran}

\end{document}